\documentclass[letterpaper, 10 pt, conference]{ieeeconf} 

\usepackage{graphicx} 
\usepackage{amsmath} 
\usepackage{subfigure}
\IEEEoverridecommandlockouts        
\title{\LARGE \bf
A Non-linear Approach to Space Dimension Perception\\ by a Naive Agent
}

\author{
Alban Laflaqui\`ere, Sylvain Argentieri, Olivia Breysse, St\'ephane Genet and Bruno Gas
\thanks{A. Laflaqui\`ere, S. Argentieri, O. Breysse, S. Genet and B. Gas are with UPMC Univ
  Paris 06 and ISIR (CNRS UMR 7222), F-75005, Paris, FRANCE
 {\tt\small name@isir.upmc.fr}}%
}

\begin{document}

\maketitle
\thispagestyle{empty}
\pagestyle{empty}

\begin{abstract}
Developmental Robotics offers a new approach to numerous AI features
that are often taken as granted. Traditionally, perception is supposed
to be an inherent capacity of the agent. Moreover, it largely relies
on models built by the system's designer. A new approach is to
consider perception as an experimentally acquired ability that is
learned exclusively through the analysis of the agent's sensorimotor
flow. Previous works, based on H.Poincar\'e's intuitions and the
sensorimotor contingencies theory, allow a simulated agent to extract
the dimension of geometrical space in which it is immersed without any
\emph{a priori} knowledge. Those results are limited to infinitesimal
movement's amplitude of the system. In this paper, a non-linear
dimension estimation method is proposed to push back this limitation.
\end{abstract}

\section{Introduction}
\label{sec:Introduction}

Traditional AI has revealed many of its limits in past decades, notably through the low adaptability of models it relies on.  Developmental Robotics attempts to overcome some of those issues by letting agents learn their own models with minimal \emph{a priori} knowledge.  One fundamental aspect of this learning process is perception as it conditions how the agent will get and analyze information. Usual methods are based on models developed in pattern recognition and signal processing inherited from Marr concepts of perception~\cite{Marr1982}. They have since been called {\it passive perception}~\cite{Bajcsy1988, Aloimonos1993} as they operate independently of the agent action on its environment.

Passive perception has been reconsidered for some time by many authors~\cite{Berthoz1997, ORegan2001a}. and new paradigms of \emph{active} perception have arisen. 
According to O'Regan and Noe {\it sensorimotor contingencies theory}~\cite{ORegan2001a}, the experience of perception would not be the activation of internal representations but the capacity to engage oneself in some structure of interaction with the environment. 
Taking inspiration from Poincar\' e's argumentation~\cite{Poincare1895} on what he called sensible space, Philipona et al.~\cite{Philipona2008} proposed a mathematical formalism to explore the sensorimotor contingencies theory. In~\cite{Philipona2004}, they describe an algorithm allowing a simple simulated agent to estimate the dimension of the geometrical space in which it is immersed without any \emph{a priori} information but its sensorimotor flow. 
 
In two recent papers~\cite{Couverture2009, Laflaquiere2010}, we proposed to take back Philipona's simulation and to prove the viability of the approach with a more bio-realistic and complex auditive modality. However, we were limited to a tangential study of the sensory manifold through unrealistic infinitesimal movements. Our goal in this paper is to exceed this linear limitation and to reach a reasonable amplitude of movement for a robot. This limitation can be overcome using non-linear projective methods, such as the Curvilinear Component Analysis (CCA)~\cite{Demartines1997}.  
Unfortunately, the non-sphericity of the distribution may prevent CCA to correctly project data in lower dimension space. Our main goal in this article is to overcome this difficulty by proposing a new active unstretching method. It mainly consists in modifying the exploration, leading to a better data distribution in terms of stretching. As such, our model is affiliated to bootstrapping methods. 

The article is divided into eight sections including this introduction and the conclusion. Section~\ref{sec:Theoretical background} is devoted to the theoretical background, giving an overview of the dimensionality estimation problem and some mathematical formalizations. Sections~\ref{sec:Simulation presentation} and~\ref{sec:Linear dimension estimation algorithm} respectively describe the simulation and the linear dimension estimation algorithm. Next section presents the dimension estimation of curved manifolds, while section~\ref{sec:Coping with non-sphericity of the data} copes with non-sphericity of the data. The article ends with a discussion in section~\ref{sec:Discussion}.

\section{Theoretical background}
\label{sec:Theoretical background}

\subsection{Poincar\'e's intuition}
\begin{figure}[h!]
\centering
\includegraphics[width=\linewidth]{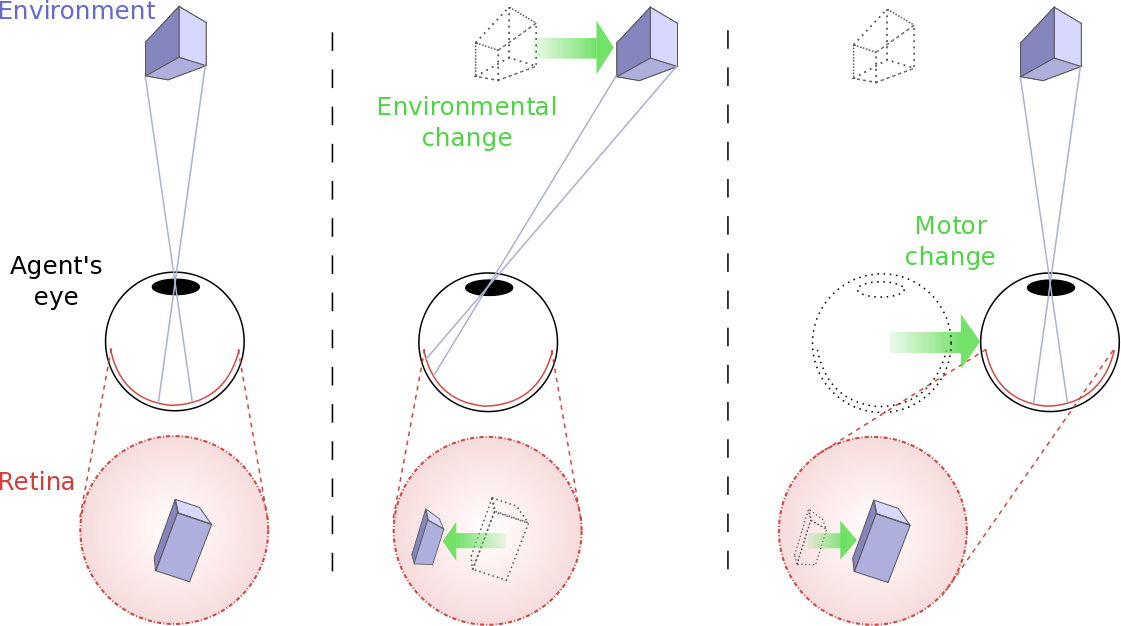}
\caption{Illustration of a compensation. After the environmental change and the motor change, the sensation on the retina is the same as in the initial state.}
\label{fig:compensation}
\end{figure}

	In 1895, H.Poincar\'e exposed his view on geometry in "L'espace et la g\'eom\'etrie"~\cite{Poincare1895}. His aim was to define geometry from the standpoint of a totally naive brain which can only have access to its sensorimotor flow. For that purpose, Poincar\'e emphasizes that the sensory space, defined by all our nervous fibers, doesn't share the homogeneity, isotropy and dimensionality properties with the \emph{geometric} space we perceive. But all of these properties can be inferred from specific sensitive changes. In particular, some sensory variations can be identified as \emph{external} and \emph{compensable}. Being external means they occur without any motor commands.
Being compensable means that they can be compensated by a motor command so that the initial sensory state, before the external and motor changes, and the final sensory state, after the external and motor changes, are identical (see Figure~\ref{fig:compensation}).
Those specific sensory variations are called \emph{displacements}. According to H.Poincar\'e, \emph{displacements} are the root of the notion of geometrical space. More precisely, the \emph{displacements} of any agent-environment system constitute a (algebraic) group whose dimension is directly related to the dimension of the geometrical space perceived by the agent.

\subsection{Mathematical formalization}
\label{sec:Mathematical formalisation}
	Poincar\'e's intuition on perception of the geometrical space's dimension has been mathematically formalized by D.Philipona in~\cite{Philipona2003}. For any sensitive and active agent in an environment, Philipona assumes that sensations generated through exploration lie on a differential manifold embedded in the (possibly high-dimensional) sensory space. Let $E$ be the environmental state (column) vector, $M$ the agent's motor state vector and $S$ the agent's sensory state vector. Then the aforementioned sensory manifold is generated through the sensorimotor law $\varphi$ so that:
\begin{equation}
S  = \varphi(C), 
\text{with}\ C  = 
\begin{pmatrix}
  M \\
 E
\end{pmatrix},
\label{eq:sensorimotor law}
\end{equation}
where $C$ denotes the system's configuration vector. Now consider an arbitrary reference sensory state $S_0=\varphi(C_0)$ together with the space of sensory \emph{variations} $dS$ around $S_0$. One can write:
\begin{equation}
  dS = \frac{\partial \varphi}{\partial M}_{|C_0}dM+\frac{\partial \varphi}{\partial E}_{|C_0}dE.
 \label{eq:decomposition}
\end{equation}
One can identify two subspaces in this equation: 
\begin{itemize}
\item The subspace of sensory variations $\{dS_{dE=0}\}$ due to motor changes only,
\item The subspace of sensory variations $\{dS_{dM=0}\}$ due to environmental changes only.
\end{itemize}
%
%
Then, the transversality property allows writing: 
\begin{equation}
\mathbf{d} = \mathbf{e}+\mathbf{m}-\mathbf{b},
 \label{eq:transversality}
\end{equation}
where $\mathbf{b}$ is the dimension of $\{dS\}$, $\mathbf{e}$ is the dimension of $\{dS_{dM=0}\}$, $\mathbf{m}$ is the dimension of $\{dS_{dE=0}\}$ and $\mathbf{d}$ is the dimension of the intersection $\{dS_{dM=0}\}\cap \{dS_{dE=0}\}$. It is fundamental to understand that the sensory variations that can be generated both through environmental or motor changes actually lie in this intersection. In other words, $\{dS_{dM=0}\}\cap \{dS_{dE=0}\}$ is the system's \emph{displacements} subspace. The dimension $\mathbf{d}$ of this intersection is thus the dimension of the algebraic group of \emph{displacements}. Note that this development is true in the vicinity of the reference point $ C_0$. Indeed, the dimensions $\mathbf{m}$, $\mathbf{e}$, $\mathbf{b}$ and $\mathbf{d}$ can vary depending on $C_0$.

To sum up, the geometrical space's dimension perceived by a naive agent can be inferred from the dimensions of $3$ sensory manifolds generated when: only the agent moves, only the environment moves and when both move.

\section{Simulation presentation}
\label{sec:Simulation presentation}
\begin{figure*}[t!]
   \centering
   \includegraphics[width=0.8\linewidth, height = 4.5cm]{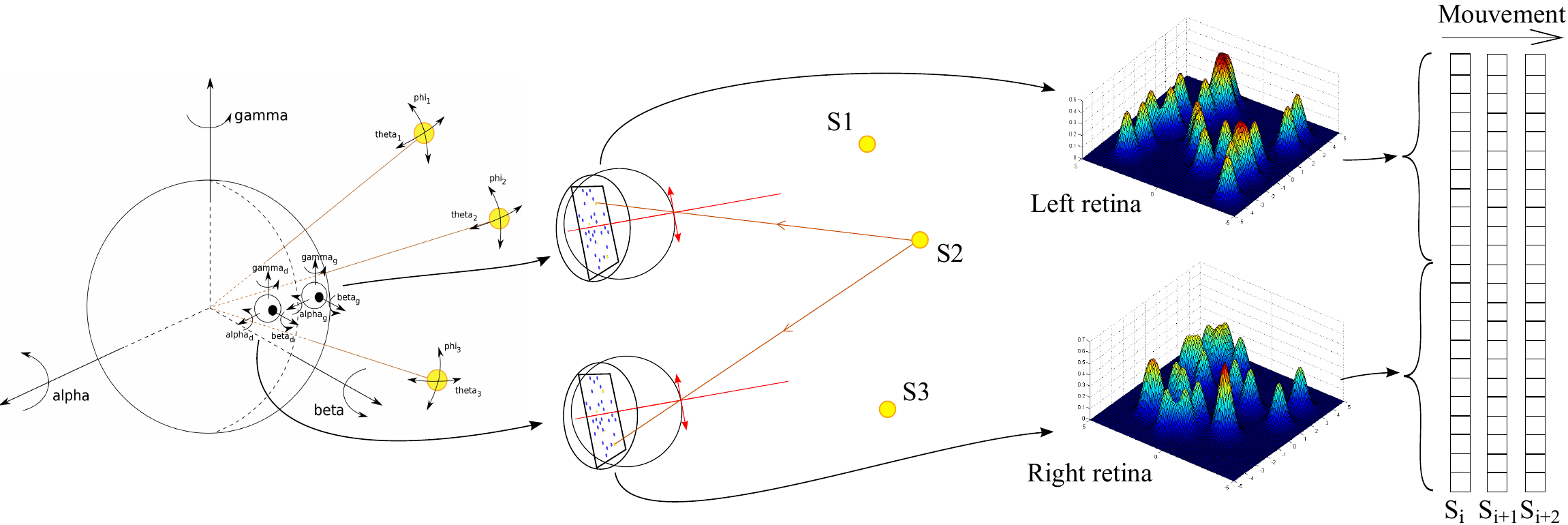}
      \caption{Schematic of the system. The agent is a head with two independent eyes. The environment is made up of souces on a 1-m sphere centered on the head. Each eye has a square retina randomly dotted with cones. Each cone is excited by the projection of sources on the retina. Finally, the sensory vector for any given configuration of the system is the concatenation of all the cones' excitations.}
   \label{fig:system}
\end{figure*}

A simple agent-environment system is proposed in order to illustrate how $\mathbf{m}$, $\mathbf{e}$ and $\mathbf{b}$ can be estimated. The whole simulation is divided into two modules (see Figure~\ref{fig:naivity}). The first part, outlined in this section, simulates the system, i.e. the interaction between the agent and its environment. The second module estimates the intrinsic dimension of the sensory manifold on the basis of the sensorimotor flow only. Methodologies for dimension estimation are described in the following sections. 
%
The simulated system consists in two subsystems immersed in a three-dimensional geometrical space (see Figure~\ref{fig:system}):
\begin{itemize}	
\item A head with two eyes, standing as the agent.
\item $N_{sources}$ punctual light sources, standing as the environment.
\end{itemize}
These are carefully described hereafter.

\subsection{Interaction modelization}
\subsubsection{The agent}
\label{sec:The agent}
The head is set up with two eyes made of a pinhole lens before a flat square retina dotted with 20 light-sensitive cells (called "cones" in the following). The reference world frame is centered on the head and aligned with the interaural frame when the head is in its initial position. Initially, the head's center is in $[0,0,0]$ cm, the left eye's pinhole is in $[-5,5,5]$ cm, the right eye's pinhole is in $[5,5,5]$ cm and the retina of each eye is $1$ cm behind its lens. The head can rotate freely along pitch ($\alpha_{h}$), roll ($\beta_{h}$) and tilt ($\gamma_{h}$) axis. Its orientation is then defined by the vector $\mathbf{r}_h=(\alpha_h,\beta_h,\gamma_h)$. Both left ($l$) and right ($r$) eyes can rotate freely and independently along pitch ($\alpha_{l}$, $\alpha_{r}$), roll ($\beta_{l}$, $\beta_{r}$) and tilt ($\gamma_{l}$, $\gamma_{r}$) axis.  Their orientations are then defined by the vectors $\mathbf{r}_l=(\alpha_l,\beta_l,\gamma_l)$ and $\mathbf{r}_r=(\alpha_r,\beta_r,\gamma_r)$. As a result, the whole agent configuration is defined by $9$ angular parameters. Each cone position on the retina is randomly drawn in $[-1,1]^2$ cm. The sensation $s_i$ generated by a cone $i$ on any retina (left or right) is:
\begin{equation}
  s_i =  \sum_{k=1}^{Nsources} \bigg( a\ \frac{\exp\big(-\textrm{dist}(cone_i,proj_k)^{2}\big)}{\textrm{dist}(eye,source_k)^{2}}\bigg), i \in \{1,20\},  
 \label{eq:vision}
\end{equation}
where $a$ is the cones sensitivity, arbitrarily set to $10^{-3}$, $\textrm{dist}(.)$ depicts the Euclidean distance, $cone_i$ is the retinal position of the $i$-th cone, $proj_k$ is the retinal position of the $k$-th source projection, $eye$ is the center of the eye in the world referential and $source_k$ is the position of the source in the world referential.

\subsubsection{The environment}
\label{sec:The environment}
The $N_{sources}$ punctual light sources lie on a $1$-meter radius sphere centered on the head. The position of each source is defined by its azimuth $\theta$ (angle between the sagittal plane and the source) and its elevation $\phi$ (angle between the transverse plane and the source). Each source can move freely and independently on the surface of the sphere. The environmental configuration is then defined by $N_{sources}\times2$ parameters.

\subsection{Simulation overview}
\label{sec:Simulation overview}

Initially, a set of $N$ configuration vectors $C_i$, \mbox{$i=\{1,\ldots,N\}$}, is randomly drawn around a working point $C_0$, with:
\begin{equation}
C_i = (\mathbf{r}_h,\mathbf{r}_l,\mathbf{r}_r,\theta_1,\ldots,\theta_{N_s},\phi_1,\ldots,\phi_{N_s})_i^T,
\label{eq:1}
\end{equation}
and stored in the matrix $\mathbf{C}=(C_1,\ldots,C_N)$. The maximal amplitude for all configuration parameters is chosen between $10^{-6}$ and $10^{+1}$ degrees. As already presented in~\ref{sec:Mathematical formalisation}, $3$ different cases can be considered: only the agent moves, only the environment moves and both move. Note that depending on the kind of exploration performed, a part of the configuration vector $C_i$ is forced to be constant and equal to its reference value. Each vector $C_i$ generates a sensory vector $S_i$ of length $n$ through the sensorimotor law $\varphi$ (see Equation \ref{eq:sensorimotor law}). Let note $\delta S_i$ the sensory variation generated thereby: 
\begin{equation}
\delta S_i=S_i-S_0=\varphi\big(C_i\big) - \varphi\big(C_0\big), \text{with } i=\{1,\ldots,N\}.
\end{equation}
The $N$ sensory variations vectors $\delta S_i$ generated by $\mathbf{C}$ are stored in the $n\times N$ data matrix $\mathbf{S} = (\delta S_1,\delta S_2,\ldots,\delta S_N)$. Finally, the data $\mathbf{S}$ is centered and reduced before being analyzed by the dimension estimation module. In the following, the dimension estimation algorithm is presented and applied to determine $\mathbf{m}$, $\mathbf{e}$, $\mathbf{b}$, and thus $\mathbf{d}$ through Equation~\eqref{eq:transversality}.

\begin{figure}[h]  
\centering 
\includegraphics[width=\linewidth]{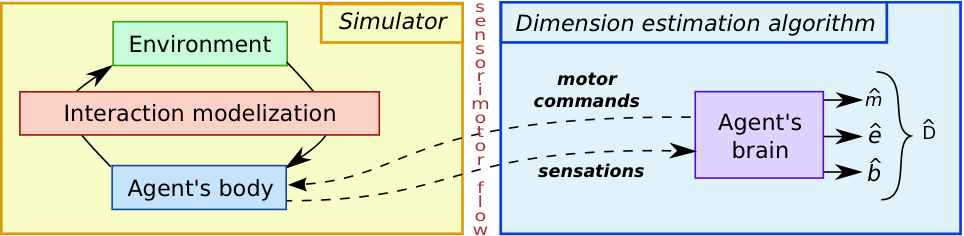} 
\caption{Schematic of the simulation.}  
\label{fig:naivity} 
\end{figure}

\subsection{Simulation parameters}
\label{sec:Simulation parameters}
In this paper, the following parameters values are exploited. For each maximal amplitude of movement in $\{10^{-6},10^{-5},10^{-4},10^{-3},10^{-2},10^{-1},10^{-0},10^{+1}\}$ degrees, these are:
\begin{itemize}
\item 3 successive simulations:  only the agent moves, only the environment moves and both move.
\item $N=1000$ movements for each exploration.
\item $20$ cones by eye, which leads to a sensory space of dimension $n=40$.
\item $N_{sources}=3$ sources in the environment.
\item $100$ successive trials of the whole process (all random parameters are redrawn) for statistical results.
\end{itemize}

\section{Linear dimension estimation algorithm}
\label{sec:Linear dimension estimation algorithm}

The second module represented in Figure~\ref{fig:naivity}, whose role is to estimate the intrinsic dimension of the data stored in $\mathbf{S}$, is described in the following. Various algorithms can be applied to perform such an estimation. A linear algorithm is first described in this section. Non-linear methods are then assessed in Section~\ref{sec:Dimension estimation of non-linear manifolds}.

\subsection{The linear algorithm}
\label{The linear algorithm}

The linear dimension estimation algorithm has already been presented in our previous paper~\cite{Laflaquiere2010}. Only the main elements are recalled in this subsection. The linear approach is based on a Singular Value Decomposition (SVD) of the data matrix $\mathbf{S}$. Let $\textbf{dim}$ be the intrinsic dimension of the manifold.
The estimated dimension $\widehat{\textbf{dim}}$ is then obtained through:
\begin{equation}
\widehat{\textbf{dim}} = \arg\max_j \Big\{\frac{\Sigma_{j}}{\Sigma_{j+1}}\Big\}, \forall j\in \big[1,\textrm{min}(n,N)-1\big],
\label{eq:ratio}
\end{equation}
where $\Sigma_j$ are the singular values of the matrix $\mathbf{S}$ in decreasing order. The estimated dimension is then equal to the number of significative singular values, as the ratio is maximal at the boundary between significative and non-significative values, see~\cite{Laflaquiere2010}.

\begin{figure*}[t!]
\centering
\subfigure[Initial data distribution.]{
   \includegraphics[width=0.23\linewidth, height = 2.9cm] {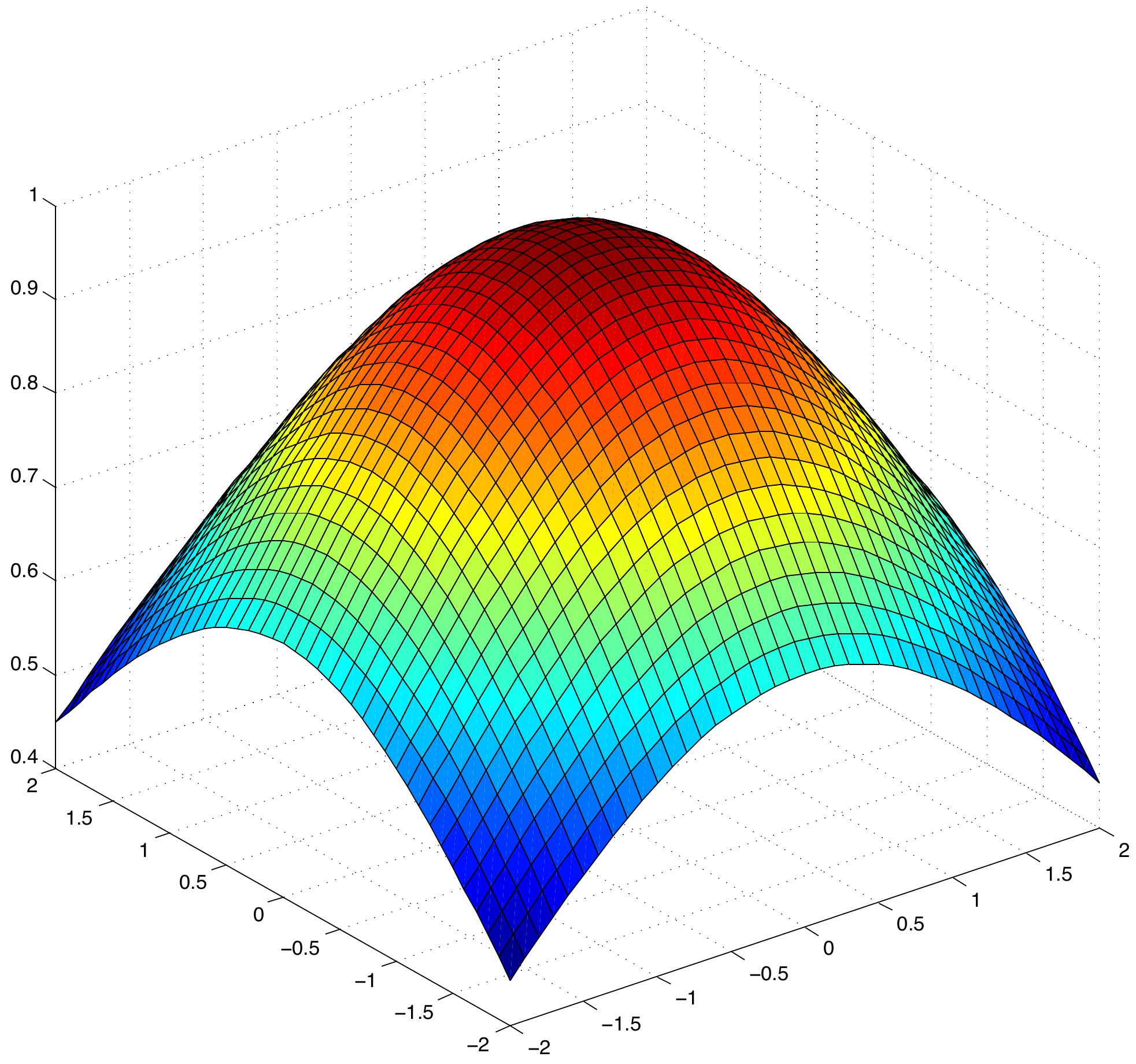}
   \label{fig:subfig1}
 }
\subfigure[Projection in a 3D space: \mbox{$J(3)=0$.}]{
   \includegraphics[width=0.23\linewidth, height = 2.9cm] {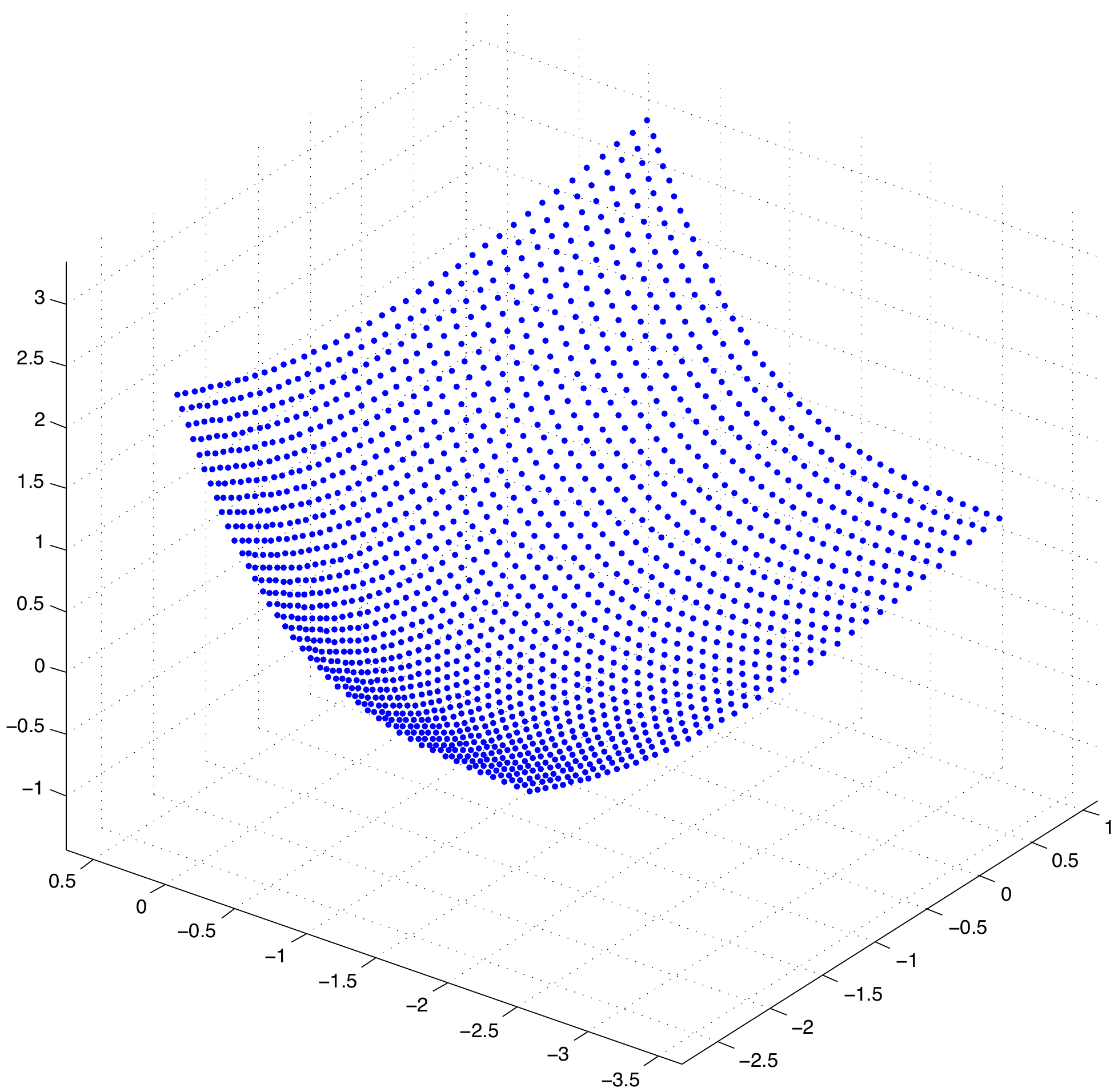}
   \label{fig:subfig2}
 }
 \subfigure[Projection in a 2D space: \mbox{$J(2)=84$.}]{
   \includegraphics[width=0.23\linewidth, height = 2.9cm] {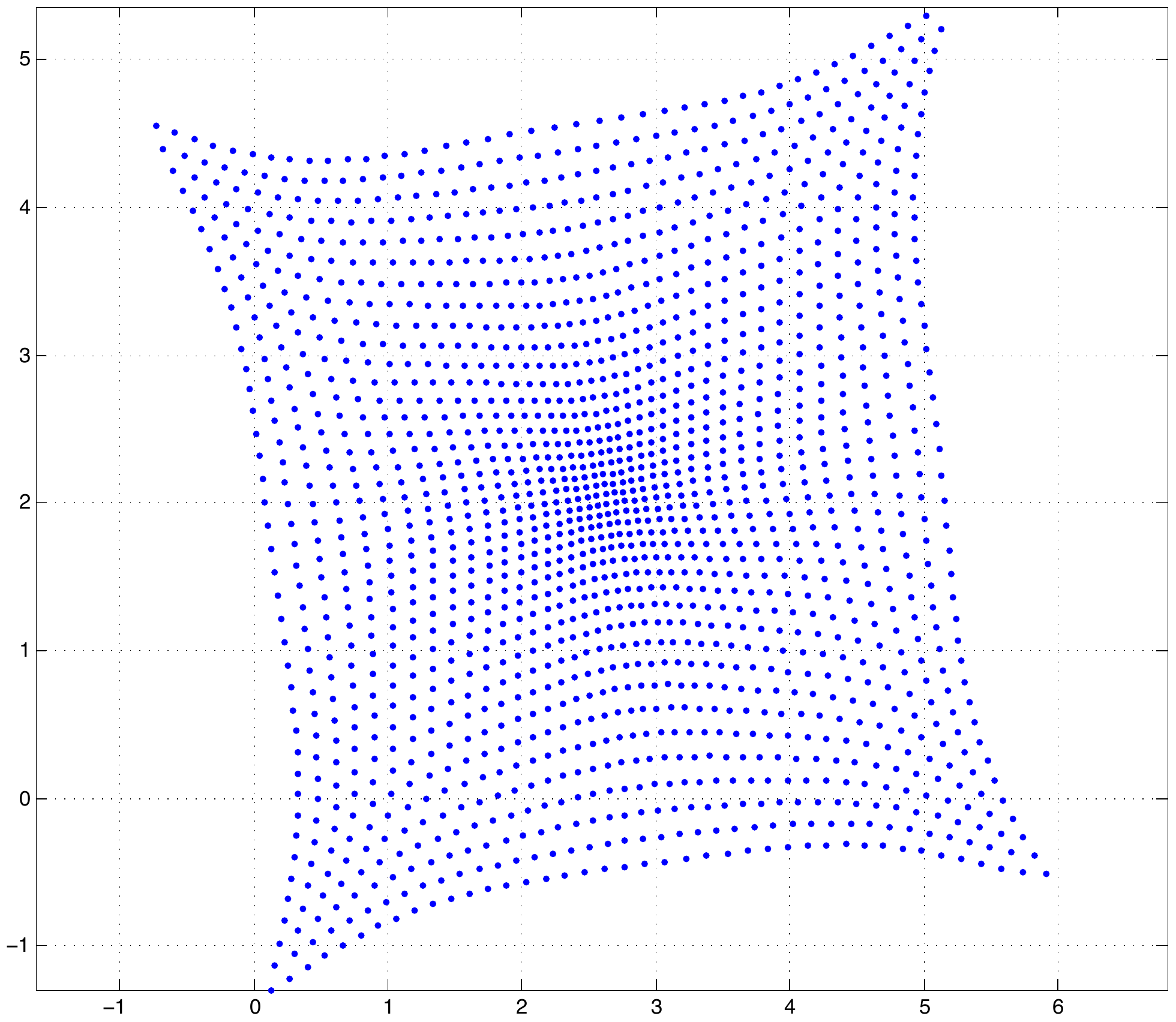}
   \label{fig:subfig3}
 }
 \subfigure[Projection in a 1D space: \mbox{$J(1)=8416$.}]{
   \includegraphics[width=0.215\linewidth, height = 2.9cm] {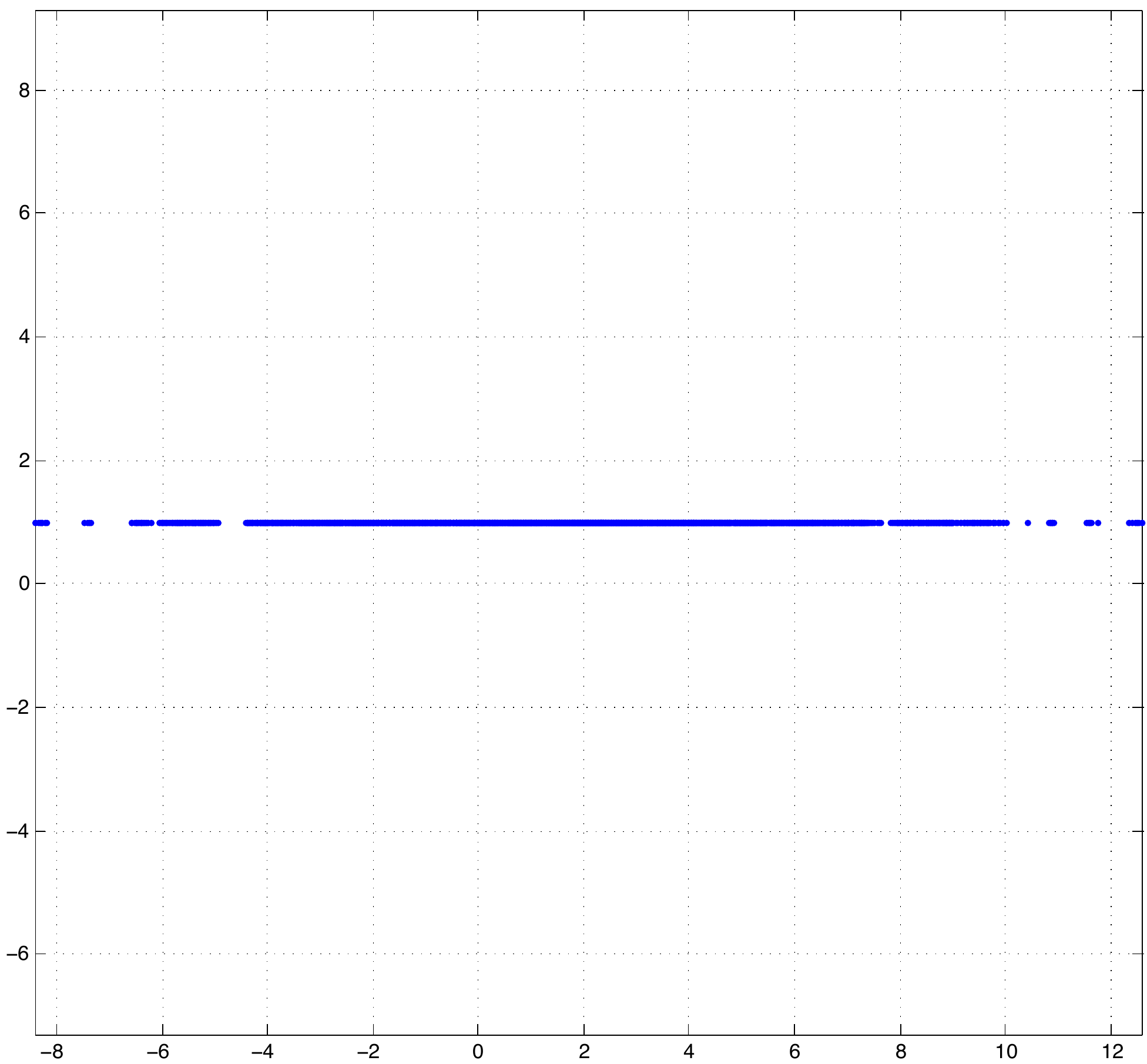}
   \label{fig:subfig4}
 }
\caption{Illustration of the CCA algorithm on a simple example. The projection of the initial data~\ref{fig:subfig1} in a 3-dimensional space~\ref{fig:subfig2} or a 2-dimensional space~\ref{fig:subfig3} preserves data topology. On the contrary, the projection in a 1-dimensional space destructs topology as there are not enough dimensions to unfold the initial manifold. As expected, the projection error $J(3)$ is null and there is a gap between $J(2)$ and $J(1)$.}
\label{fig:projection}
\end{figure*}

\subsection{Results with the linear method}
\label{sec:Results with the linear method}

The performance of the linear dimension estimation method in relation to the maximal amplitude of movement is presented in this subsection. Performances are defined as the percentage of correct dimension estimations over the $100$ successive trials.
%
The correct intrinsic dimension $\mathbf{m}$ is $9$ as the agent has $9$ degrees of freedom (see~\ref{sec:The agent}). It is $6$ for $\mathbf{e}$ as each of the $3$ light sources has $2$ degrees of freedom (see~\ref{sec:The environment}). Finally, as designer of the system, we know that the $3$ rotations centered on the head are its only compensable movements. The dimension $\mathbf{d}$ of the \emph{displacements}' algebraic group is then $3$. (Note that if translations of the head and of the sources were allowed, this dimension would be $6$, as they would be compensable too). According to Equation~\eqref{eq:transversality}, the correct intrinsic dimension $\mathbf{b}$ is then $12$.  Figure~\ref{fig:performances} shows that the performance of the linear method is good for a maximal movement amplitude set to $10^{-6}$ degrees, but drops progressively for greater magnitudes. This can be explained by the increasing curvature of the data distribution when movement amplitude increases~\cite{Laflaquiere2010}. One can also notice that the drop is slower in the $\mathbf{e}$-case (i.e.\ when only the environment moves). This might be due to a lower curvature of the data distribution than in the $2$ other types of exploration.

The problem faced when considering non-infinitesimal amplitude of movement in the system is the non-linearity of the sensorimotor law $\varphi$, which might lead to curved sensory manifolds. The linear method is then unsuitable when considering realistic amplitudes of movement for a robotic or biologic agent. In the following, a non-linear dimension estimation method will be selected to cope with this curvature issue.

\section{Dimension estimation of non-linear manifolds}
\label{sec:Dimension estimation of non-linear manifolds}

Dimension estimation of a manifold is a well formalized and an easily solved problem when its curvature is null. However, this estimation is still challenging when dealing with curved manifolds in a high-dimensional space. For more information about non-linear methods, a comparative overview can be found in \cite{Lee2007,Tsai2010}.

\subsection{Non-linear method}
\label{sec:Non-linear method}

In all the following, the dimension estimation relies on a projection method: the Curvilinear Component Analysis (CCA)~\cite{Demartines1997}. Indeed, it has been experimentally proven that CCA gives better results in high-dimensional spaces than other methods, even if the manifold is highly curved~\cite{Lee2007}. Moreover, its complexity is $O(N)$ whereas other equivalent methods present a $O(N^2)$ complexity. 

CCA is an iterative method projecting data from a $n$-dimensional to a $p$-dimensional space, with $p\leq n$. For that purpose, a self-organized neural network is exploited to minimize a cost function $J(p)$ assuring the conservation of the data topology through local distances preservation. More information about the algorithm can be found in~\cite{Demartines1997}.
The value of $J(p)$ at the last iteration represents how well the local distances have been respected when projecting the data.

In all the following, the initial and final learning rates are set to $5.10^{-1}$ and $5.10^{-4}$, the initial and final neighborhood are set to $4$ and $0.2$, the number of points is set to $N=1000$ and the number of iterations to $K=100$.

\subsection{Non-linear dimension estimation}

The non-linear dimension estimation method is based on the computation of the cost function $J(p)$, with $p \in [1,15]$. Note that $\max(p)$ is set to $15$ so as to limit the high computational cost of the whole process. The projection error $J(p)$ is small when $p\ge\textbf{dim}$. Indeed, in this case, there are enough embedding dimensions to unfold the manifold while preserving its topology. On the contrary, the projection error becomes significant when $p<\textbf{dim}$, as the manifold unfolding is performed to the detriment of its topology conservation. So, the ratios of successive projection costs exhibit a maximum at the boundary between significative and non-significative projection costs values. Consequently, the estimated intrinsic dimension of the dataset $\widehat{\textbf{dim}}$ is then defined as:
\begin{equation}
\widehat{\textbf{dim}}=\arg\max_p \Big\{\frac{J(p-1)}{J(p)}\Big\}, \forall p\in [2,15].
 \label{eq:estimation}
\end{equation}

Figure~\ref{fig:projection} illustrates the principle of the dimension estimation algorithm based on CCA on a simple distribution. In this example, a 2D manifold embedded in a 3D space is successively projected on 3D, 2D and 1D spaces. For each projection, the cost function $J(p)$ is provided, exhibiting its correlation with the topology conservation. 

\subsection{Results with the non-linear method}
\label{Results with CCA}
The aforementioned CCA algorithm is now tested on the data originating from the simulated system presented in \S\ref{sec:Simulation presentation}. In the following, the simulation parameters are the same as in part~\ref{sec:Simulation parameters}, and the number of CCA iterations is set to $K=100$. Results are reported on Figure~\ref{fig:performances}. It shows that the CCA is not able to precisely estimate each dimension. More precisely, $\mathbf{m}$ and $\mathbf{b}$ are never correctly estimated, while the performance for $\mathbf{e}$ is only around $40\%$ for all amplitudes. But while the overall performance of the method is surprisingly low, 
one can notice that it does not vary with the movement amplitude. This seems to indicate that the curvature is not at the origin of the bad estimations, which are more likely caused by the data distribution properties. This hypothesis will be investigated in the next section.

\section{A bootstrap approach to the dimension estimation problem}
\label{sec:Coping with non-sphericity of the data}

Among all possible data distribution properties, curvature and data density are known to be well addressed by CCA~\cite{Lee2007}. Nevertheless, the non-sphericity of the data distribution can be a major cause of failure for the algorithm. The notion of sphericity, similar to the stretching of the dataset, is first discussed. Then, an iterative unstretching approach is proposed. Finally, this method is combined with CCA to provide a new dimension estimation technique.

\subsection{Stretching of the data}
\label{sec:Stretching of the data}
\begin{figure}
\centering
\includegraphics[width=0.7\linewidth, height = 3cm]{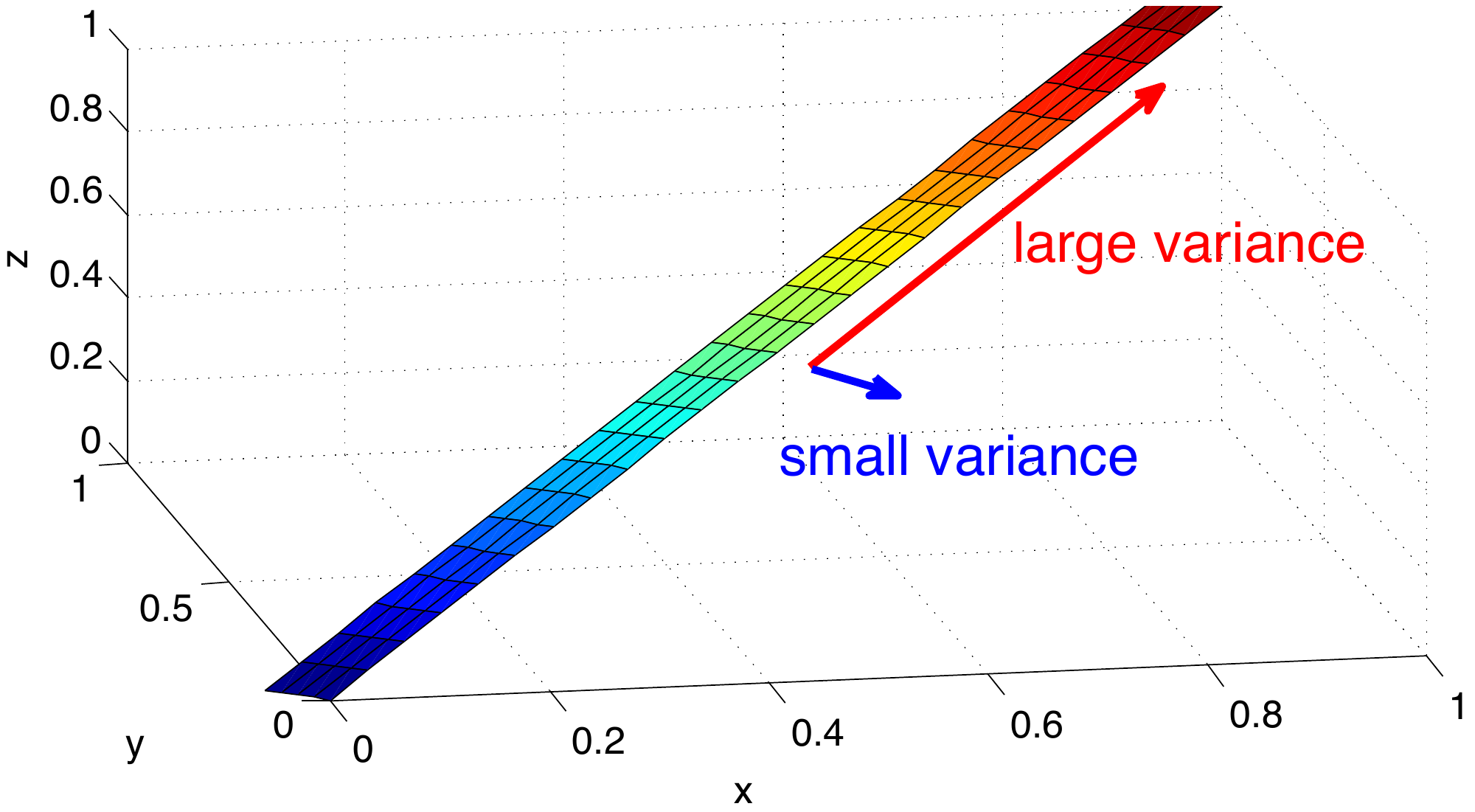}
\caption{Example of a 2D plane manifold in a 3D space}
\label{fig:stretching}
\end{figure}	
In order to understand the notion of non-sphericity,  a simple illustration is provided in Figure~\ref{fig:stretching}. In this example, a noise-free 2D linear dataset is stretched in a 3D hypercube. In this case, the diagonal explains a very large part of the data variance while other orthogonal directions support very few or none. If the stretching is too strong, every non-linear dimension estimation algorithm will fail in finding the right intrinsic dimension of the distribution ($\textbf{dim}=2$) and will instead find $\widehat{\textbf{dim}}=1$. Even though one can argue that this result is correct --because all directions orthogonal to the diagonal can be considered as noise--, a linear algorithm would compute the correct intrinsic dimension $2$ as this manifold is flat. This poor performance of non-linear algorithms in the stretched case is due to their inherent tolerance to changes along axis with small variance. 
For instance, the projection of the 2D manifold in Figure~\ref{fig:stretching} would lead to projection errors $J(2)$ and $J(1)$ being both very small, while $J(1)$ was expected to be significantly greater than $J(2)$, like in Figure~\ref{fig:projection}.

To insure an accurate estimation of the manifold's dimension, the strongly-stretched case must be avoided. The preliminary centering and reduction of the data doesn't solve the problem. Indeed the reduction may have no impact on the manifold's shape, as illustrated in the previous example (see Figure~\ref{fig:stretching}) where the data already spans the same range on each axis of the 3D space. A better solution is to perform a PCA on the dataset and to normalize its variance along eigenvectors. Note that this process makes sense only for linear or very slightly curved manifolds. However, this approach has one major drawback: the normalization can lead to a huge noise amplification if the data's variance along some axis is similar or smaller than noise's variance. In order to avoid this problematic noise amplification, an active unstretching method is proposed in the next subsection.
	
\subsection{The bootstrapping}
\label{sec:The bootstrapping}

\subsubsection{Relevance of an active unstretching}
The shortcoming of the PCA/reduction approach is to consider only the final sensory data, but not the way it has been generated. During the reduction, noise is then likely to be reshaped, reduced or amplified along with the data. On the contrary, generating new data by modifying the system's movements can lead to a better exploration of the manifold (in terms of stretching), while maintaining a constant level of noise. 
The goal of the unstretching method is to modify the system's movements so that the data presents a similar variance along the $\textbf{dim}$ axis underlying the manifold. Of course, it doesn't affect its dimension $\textbf{dim}$ which only dependents of the agent-environment interaction's properties. The way the system moves regards only the portion of the whole sensory manifold actually explored during the experiment.

\subsubsection{Presentation of the bootstrapping}	
	The original active unstretching approach outlined in this paper is an iterative method of resampling (bootstrap) based on a SVD. Let $\mathbf{C^{(b)}}$ be the configurations matrix (defined in~\S\ref{sec:Simulation overview}) at the $b$-th bootstrapping iteration. As already mentioned, a data set $\mathbf{S}$ is generated by $\mathbf{C^{(b)}}$ through the sensorimotor law $\varphi$ (see Equation~\ref{eq:sensorimotor law}). The singular value decomposition of $\mathbf{S}$ provides a $n \times n$ unitary matrix $\mathbf{U}$, a  $N \times N$ unitary matrix $\mathbf{V}$ and a $n \times N$ diagonal matrix $\mathbf{\Sigma}$ with non-negative real numbers on the diagonal, such that: 
\begin{equation}
\mathbf{\Sigma}=\mathbf{U}\mathbf{S}\mathbf{V^T}.
 \label{eq:SVD}
\end{equation}	
The singular values $\Sigma_{j}$ of $\mathbf{S}$ are sorted in decreasing order on the diagonal of $\mathbf{\Sigma}$, with $j = \{1,\ldots,\min(n,N)\}$. The right singular vectors stored in the columns of $\mathbf{V}$ are linear combinations of the $N$ command vectors of  $\mathbf{C^{(b)}}$. The first $\min(n,N)$ of them are directly related to the singular values $\Sigma_{j}$. They form a base $\mathcal{V}$ in which the variability of $\mathbf{S}$ can be explained.

The bootstrapping method consists in a modification of the exploratory commands $\mathbf{C^{(b)}}$ so that the singular values $\Sigma_{j}$ tend to be identical at the iteration $b+1$. Moreover, the maximal amplitude of movement allowed must remain constant. Finally, the algorithm is made of the following steps for iteration $b\leq B$:

\begin{itemize}
\item $\mathbf{S}$ is obtained through \mbox{$\mathbf{S} = \varphi(\mathbf{C^{(b)}})$};

\item $\mathbf{S}$ and $\mathbf{C^{(b)}}$ are normalized by their maximum value as absolute values are meaningless in following steps and $\max(|\mathbf{C^{(b)}}|)$ is denoted $Cmax$;

\item A SVD is applied to $\mathbf{S}$, providing the matrix $\mathbf{V}$ from which is extracted $\mathcal{V}$;

\item $\mathcal{V}$ is expressed in the initial base of dof (the degrees of freedom exhibited in equation \ref{eq:1}):
 \begin{equation}
 \mathcal{V}'=\mathbf{C^{(b)}}\mathcal{V};
  \label{eq:changement de base 1}
\end{equation}
The column vectors of $\mathcal{V}'$ may contain useless components that have no influence on the sensory variation around the working point. They could be either \emph{displacement} or redundancy of the command system. They are invisible for the bootstrap as its inputs are sensory variations, but they can still be amplified as its outputs are system commands. To avoid their uncontrolled amplification, those components are removed from $\mathcal{V}'$ using the $4$ following steps.
\item The remaining vectors of $\mathbf{V}$ are expressed in the initial base of dof:
 \begin{equation}
 \mathcal{V}_2'=\mathbf{C^{(b)}}V\Big(1\ldots N,\big(\min(n,N)+1\big)\ldots N\Big);
  \label{eq:changement de base 2}
\end{equation} 
The column vectors of $\mathcal{V}_2'$ cover the part of the dof space that do not generate sensory variations.
\item A SVD is performed on $\mathcal{V}_2'$, providing $[\mathbf{U^2},\mathbf{\Sigma^2},\mathbf{V^2}]$ such that $\mathbf{\Sigma^2}=\mathbf{U^2}\mathcal{V}_2'(\mathbf{V}^{2})^{T}$.
\item The singular values $\Sigma^{2}_{j}$ greater than a threshold of $1$ are considered significative. Their related left singular vectors $U^2_{j}$ form the base $\mathcal{U}$ of commands that don't generate sensory variations.
\item Useless components are removed from $\mathcal{V'}$:
 \begin{equation}
 \mathcal{V}' = \mathcal{V}' - \mathcal{U}\mathcal{U}^+\mathcal{V}'
  \label{eq:retire compensables}
\end{equation}
where $^+$ depicts the pseudoinverse.
\item $\mathbf{C^{(b)}}$ is expressed in the base $\mathcal{V}'$:
\begin{equation}
\mathbf{C^{(b)}}'=(\mathcal{V}')^+\,\mathbf{C^{(b)}},
\label{eq:changement de base 3}
\end{equation}
Entries of the command vectors of $\mathbf{C^{(b)}}'$ are then directly related to the singular values $\Sigma_{j}$.
\item Vectors of $\mathbf{C^{(b)}} '$ are reshaped: 
\begin{equation}
\mathbf{\mathbf{C^{(b+1)}}}'=\mathbf{\Gamma}\mathbf{C^{(b)}} ',
\label{eq:reshaping}
\end{equation}
where $\mathbf{\Gamma}$ is a diagonal matrix defined by:
\begin{align}
\text{diag}(\mathbf{\Gamma}) & = [\Gamma_1, \ldots, \Gamma_{\min(n,N)}], \text{ with}\\
\Gamma_j & = \textrm{min}\Big\{\textrm{ln}(\Sigma_{1}/\Sigma_{j})+1,L\Big\}.
\label{eq:coefficients}
\end{align}
Command vectors are thereby amplified along the directions that previously generated the lowest variance in the sensory space. The logarithmic function and the threshold coefficient $L$ limit the influence of high ratios due to strong initial stretching or noise which can lead to very small $\Sigma_{j}$ values. In this implementation, $L$ is arbitrary fixed to $10$.
\item Vectors of $\mathbf{C^{(b+1)}}'$ are expressed in the initial configuration base: 
\begin{equation}
\mathbf{C^{(b+1)}}=\mathcal{V}'\mathbf{C^{(b+1)}}'.
\label{eq:changement de base 2}
\end{equation}
\item Vectors of $\mathbf{C^{(b+1)}}$ are normalized:
\begin{equation}
\mathbf{C^{(b+1)}}=\frac{Cmax}{\max(|\mathbf{C^{(b+1)}}|)}\,\mathbf{C^{(b+1)}}.
\label{eq:normalisation}
\end{equation}
The maximal amplitude for a configuration parameter is then identical before and after the iteration. Without, this step, amplitudes would diverge as $\mathbf{C^{(b)}} '$ reshaping has been done through amplification of entries.
\item The whole process is repeated until convergence of the singular values of $\mathbf{S}$, or for a given number of iterations $B$.
 \end{itemize}

\subsubsection{Discussion on the bootstrap algorithm}
\label{sec:Remarks on the boostraping}

Strictly speaking, the whole aforementioned algorithm only makes sense for linear or slightly curved manifolds, as it is based on a SVD of the data matrix. Two strategies are then proposed to integrate it in the exploration strategy:
\begin{itemize}
\item  In order to respect the linearity hypothesis, the bootstrapping is performed with an infinitesimal movement amplitude (typically $10^{-6}$ degrees). A last exploration is then executed with a command set $C^{(B+1)}$ which corresponds to $C^{(B)}$ amplified to reach the desired movement amplitude.
\item The bootstrapping is directly performed with the desired movements amplitude.
\end{itemize}
In the first case, the behavior and convergence of the boostraping is guaranteed but the influence of greater commands on the data distribution is not taken into account.
In the second case, no infinitesimal amplitude is necessary but the good behavior of the bootstrapping cannot be guaranteed. For example, curvature can prevent the algorithm to stretch the data variance along one axis without affecting other ones. Consequently, singular values can converge toward distinct magnitudes instead of a common one during the bootstrapping. This approach may however be sufficient to reach more realistic movement magnitudes, thus allowing the implementation on a real robotic platform.

\subsection{Results of the non-linear method with bootstrapping}
\label{sec:Results of CCA with bootstrapping}

The performance of the CCA method with the $2$ bootstrapping strategies, in relation to the maximal amplitude of movement, is discussed in this section. For each maximal amplitude of movement, the simulation parameters are the same as in \S\ref{sec:Simulation parameters}. Moreover, the number of bootstrap iterations is set to $B=10$.

\begin{figure}[b]
   \centering
   \includegraphics[width=\linewidth, height=7cm]{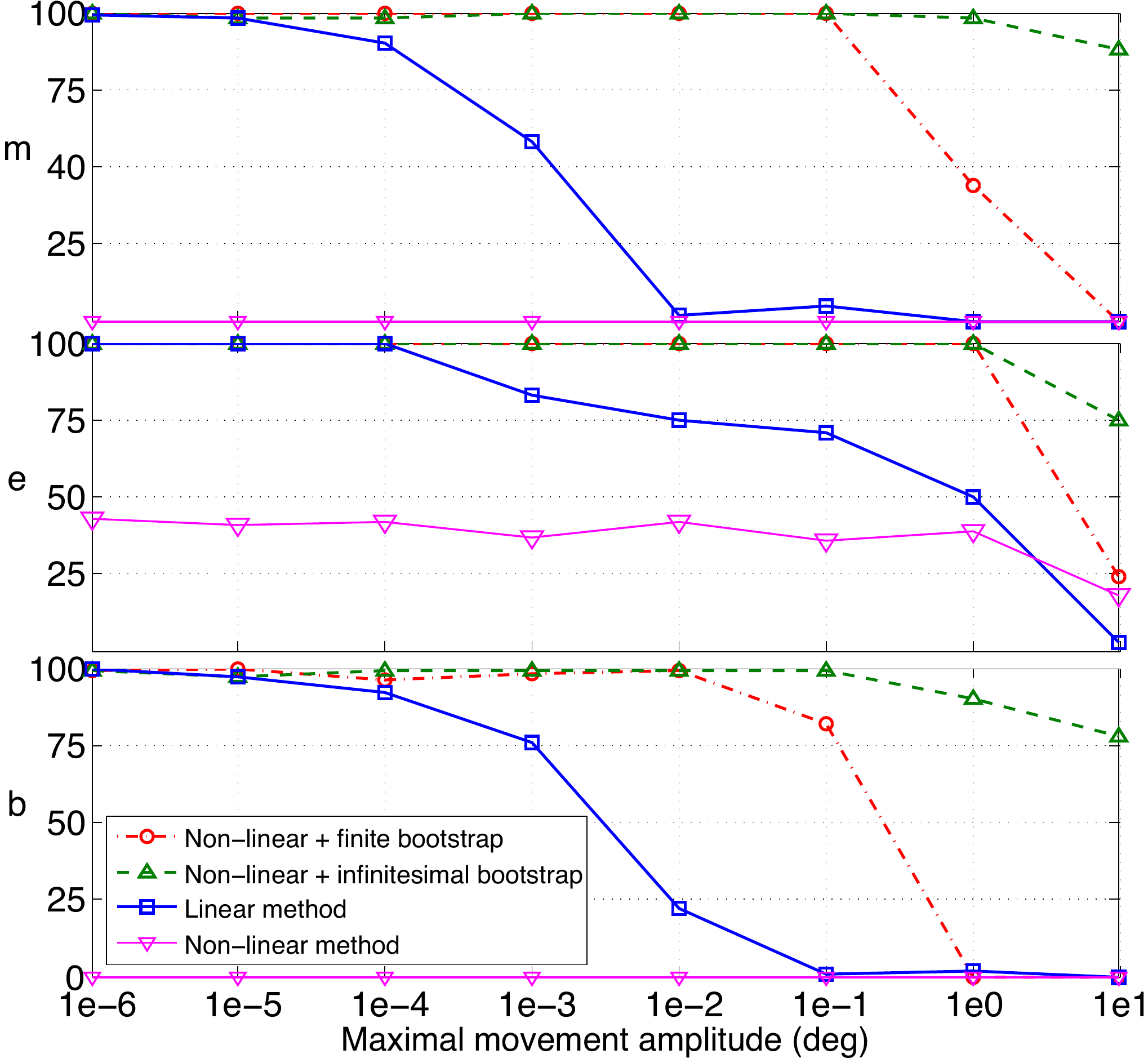}
      \caption{Performance of the 4 methods for the estimation of $\mathbf{m}$, $\mathbf{e}$ and $\mathbf{b}$.}
   \label{fig:performances}
\end{figure}

Figure~\ref{fig:performances} shows that the performances of the non-linear method with boostraping are better than with the two previous  methods.
It shows that, with a more symmetric distribution, CCA can handle curvature as its performance is satisfactory up to amplitude $10^{-1}$ degrees ($10^{0}$ degrees for $\mathbf{e}$), where the linear method performance is already very poor or null. Between the $2$ bootstrapping strategies, the infinitesimal one has better performance for the greatest amplitudes. This result was expected as this strategy guarantee the good behavior of the bootstrapping for any amplitude. On the contrary, the second strategy doesn't seem to converge to interesting solutions when movements amplitude, and so curvature, becomes too important.
However, even with the infinitesimal bootstrap, performance decreases for amplitudes greater than $10^{-1}$ degrees. This slight drop could be due to unexpected data distribution that is not taken into account while performing the infinitesimal bootstrap, but more probably to CCA that needs an increasing number of iteration and points when the manifold curvature grows.


\section{Discussion}
\label{sec:Discussion}

The non-linearity of the sensorimotor law $\varphi$ leads to curved sensory manifolds, making linear methods inadequate when dealing with realistic movement amplitudes. 
Moreover, the basic non-linear method shows poor estimation performance because of the stretching of the data distribution. With the active resampling process the non-linear method based on CCA is able to estimate dimensions $\mathbf{m}$, $\mathbf{e}$ and $\mathbf{b}$. However, their performances decrease for too great amplitudes. This drop is due to the limitation of the underlying linear hypothesis when considering the bootstrap with finite amplitudes. For the infinitesimal bootstrap strategy, it might be due to unexpected data distribution, but more probably to a limitation of CCA which needs more iterations when the manifolds curvature increases.

The non-linear method proposed in section~\ref{sec:Coping with non-sphericity of the data} pushes back the linear limitation introduced in~\cite{Philipona2004}. However, the best results are obtained with the infinitesimal bootstrapping method which doesn't respect the goal of finite movement amplitude. The finite amplitude bootstrapping exhibits correct performances up to movements of $10^{-1}$ degrees, which can be considered as a reasonable amplitude for robotic systems and make conceivable an implementation on a real platform.
Nevertheless, even greater or unlimited amplitudes would be an interesting objective to validate Poincar\'e's intuition, specially for biological agents. To do so, the bootstrapping process should be replaced by a non-linear intelligent exploration strategy to guarantee that the data distribution presents symmetry properties good enough for any non-linear dimension estimation method. A dynamical study of the whole system, instead of the static study presented in this paper, could also be a way to surpass the current limitations.

\section{Conclusion}
\label{sec:Conclusion}

An active and model-free feature extraction approach from a high-dimensional dataset has been presented in this paper. It is based on the idea that the perception can be considered as an experimentally acquired ability, exclusively learned through the analysis of an agent's sensorimotor flow. This purpose is illustrated with a simple simulated system, in which the agent is able to estimate the dimension of the geometrical space in which it is immersed. This is made possible thanks to the use of an original data analysis approach, relying on a non-linear dimension estimation algorithm mixed with an active resampling strategy. Results show that, contrarily to previous linear approaches, realistic movement magnitudes are now accessible to the agent. This will allow us to work in a close future on the validation of this theory on real robotics platforms.
Finally, this work opens up a new way to consider space perception. The estimation of its dimension is only a preliminary step towards the structuring of space from the point of view of the active agent. The base $\mathcal{V}$ (see equation~\ref{eq:SVD}) already provides us a way to move in the \emph{displacements} space and could lead in future work to goal oriented behaviors of the agent. At last, the sensorimotor analysis approach used in this paper could be applied to other fields of perception. The features extracted with such tools would be intrinsically meaningful for the agent and could lead to new IA algorithms.

\section*{Acknowledgment}
This work was conducted within the French/Japan BINAAHR (BINaural Active Audition for Humanoid Robots) project under Contract n$^\circ$ \mbox{ANR-09-BLAN-0370-02} funded by the French National Research Agency.

\bibliographystyle{plain}
\bibliography{iros12.bib}
\end{document}